\documentclass[11pt]{article}
\usepackage[preprint]{acl}

\usepackage{times}
\usepackage{latexsym}
\usepackage{booktabs}
\usepackage{pdflscape}
\usepackage{makecell}
\usepackage{booktabs}  
\usepackage{siunitx}   
\usepackage{enumitem}
\usepackage{multirow}
\usepackage{subcaption}
\usepackage{graphicx}
\usepackage{tabularx}
\usepackage{xcolor}
\usepackage{colortbl}
\usepackage{arydshln}
\usepackage{fontawesome5}
\usepackage{float}
\usepackage{tabularx}
\usepackage{pifont}
\newcommand{\cmark}{\ding{51}}%
\newcommand{\xmark}{\ding{55}}%
\usepackage{footmisc}

\usepackage[T1]{fontenc}

\usepackage{microtype}

\usepackage{inconsolata}
\usepackage{hyperref}

\usepackage{graphicx}

\newcommand{\data}{\textsc{MMCricBench-3K}}
\newcommand{\datah}{\textsc{MMCricBench-H-1.5K}}
\newcommand{\datae}{\textsc{MMCricBench-E-1.5K}}
\title{Mind the (Language) Gap: Towards Probing Numerical and Cross-Lingual Limits of LVLMs}

\author{Somraj Gautam\thanks{Equal contribution.}, Abhirama Subramanyam Penamakuri\footnotemark[1], Abhishek Bhandari and Gaurav Harit \\
  Indian Institute of Technology Jodhpur\\
  \texttt{\{gautam.8,penamakuri.1,bhandari.1,gharit\}@iitj.ac.in}\\
  \href{https://huggingface.co/datasets/DIALab/MMCricBench}{\textbf{{https://huggingface.co/datasets/DIALab/MMCricBench}}}}

\begin{document}
\maketitle

\begin{abstract}

We introduce \data{}, a benchmark for Visual Question Answering (VQA) on cricket scorecards, designed to evaluate large vision-language models (LVLMs) on complex numerical and cross-lingual reasoning over semi-structured tabular images. \data{} comprises 1,463 synthetically generated scorecard images from ODI, T20, and Test formats, accompanied by 1,500 English QA pairs. It includes two subsets: \datae{}, featuring English scorecards, and \datah{}, containing visually similar Hindi scorecards, with all questions and answers kept in English to enable controlled cross-script evaluation. The task demands reasoning over structured numerical data, multi-image context, and implicit domain knowledge. Empirical results show that even state-of-the-art LVLMs, such as GPT-4o and Qwen2.5VL, struggle on the English subset despite it being their primary training language and exhibit a further drop in performance on the Hindi subset. This reveals key limitations in structure-aware visual text understanding, numerical reasoning, and cross-lingual generalization. The dataset is publicly available via Hugging Face at \url{https://huggingface.co/datasets/DIALab/MMCricBench}, to promote LVLM research in this direction.

\end{abstract}
\begin{figure}[ht]
\centering
\includegraphics[width=.48\textwidth]{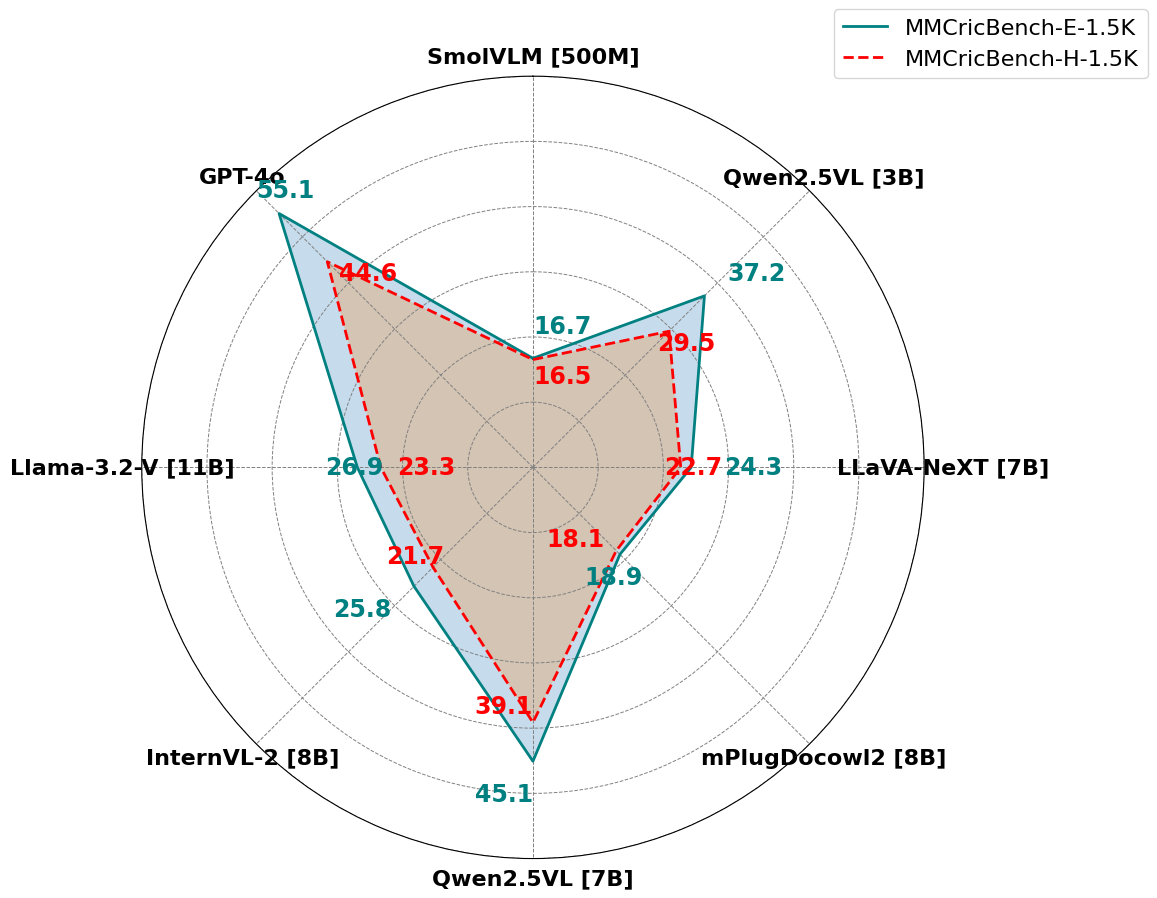}
\caption{LVLM performance on \datae{} (English) and \datah{} (Hindi) cricket scorecards. While accuracy on English scorecards peaks at 55.1\%, performance on visually similar Hindi scorecards remains consistently lower, highlighting a persistent gap in cross-lingual structure-aware numerical reasoning over images.}
\label{fig:Spider_chart}
\end{figure}

\begin{figure*}[ht]
\centering
\includegraphics[width=\textwidth]{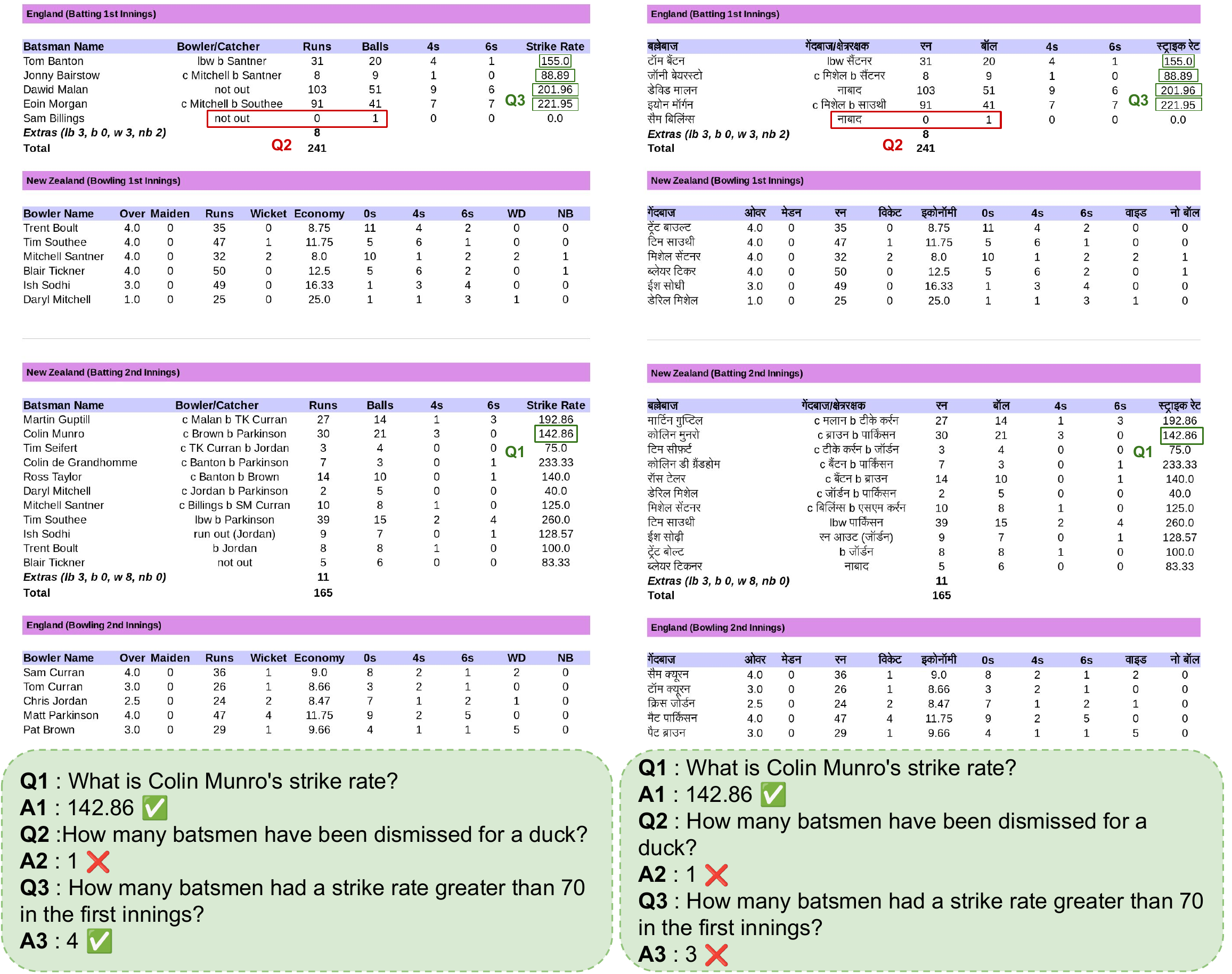}
\caption{Examples of LVLMs (dis)parity between \datae{} and \datah{}. Example predictions by Qwen2.5VL-7B on English (left) and Hindi (right) scorecards. Q1 is a simple retrieval question, correctly answered in both cases. Q2 requires structure-aware, domain-specific reasoning, leading to failure in both. Q3 reveals a cross-lingual gap answered correctly on the English scorecard but incorrectly on the Hindi one, despite identical content.}
\label{fig:first_diagram}
\end{figure*}

\section{Introduction}
Text-centric visual question answering (VQA) has seen considerable progress with benchmarks such as TextVQA~\cite{singh2019towards}, ST-VQA~\cite{xia2023st}, DocVQA~\cite{mathew2021docvqa},  VisualMRC~\cite{tanaka2021visualmrc}, and OCRBench~\cite{liu2024ocrbench}, which evaluate models on tasks requiring OCR-based understanding and textual reasoning. More recently, tabular VQA datasets like TableVQA-Bench~\cite{kim2024tablevqa}, TabComp~\cite{gautam2025tabcomp}, and ComTQA~\cite{zhao2024tabpedia} have introduced structure-aware challenges focusing on numerical reasoning and table comprehension. However, as summarized in Table~\ref{tab:statistics2}, these benchmarks often fall short in one or more dimensions: they are primarily monolingual (mostly English), lack multi-image contextual reasoning, and offer limited evaluation of fine-grained domain-specific numerical reasoning.

Cricket scorecard images, on the other hand, represent a compelling testbed for evaluating such capabilities. These semi-structured layouts combine tabular numeric data (runs, overs, wickets) with implicit contextual information (e.g., Which bowler has bowled the most wides in the match? Q3 in Figure~\ref{fig:first_diagram}), sometimes spanning across multiple images. In this work, we introduce {\data{}}, a novel benchmark for visual question answering on cricket scorecards, designed to evaluate the \textit{structure-aware}, \textit{mathematical}, \textit{multi-image}, and \textit{cross-lingual} reasoning capabilities of large vision-language models (LVLMs). \data{} comprises 1,463 synthetically generated scorecard images (822 single-image and 641 multi-image examples), along with 1,500 English QA pairs. It includes two subsets: {\datae{}} (English scorecards) and {\datah{}} (Hindi scorecards), with all questions and answers provided in English to enable controlled evaluation across script variations.

Large Vision-Language Models (LVLMs) (LLaVA-1.5~\cite{liu2024visual}, MiniGPT4~\cite{chen2023minigpt}, mPLUG-Owl~\cite{ye2024mplug}, Qwen-VL~\cite{wang2024qwen2}, and InternVL2~\cite{chen2024internvl}) have become the de facto approaches for visual question answering tasks, including text-aware visual tasks. Recent LVLMs such as Qwen2.5VL~\cite{bai2025qwen2}, mPLUG-DocOwl2~\cite{hu2024mplugdocowl2}, InternVL2~\cite{chen2024internvl}, and TextMonkey~\cite{liu2024textmonkey} have further advanced the bar on text-aware tasks, including VQA, by incorporating text-aware objectives into their pretraining or instruction-tuning stages. While studies exist to show strong performance of these models on English benchmarks, similar studies to understand their robustness across low-resource languages like Hindi\footnote{Hindi as a textual language is not a low-resource language, however, when we look at visual text space, Hindi is a low-resource language.} remains unexplored in the literature. 
To this end, we leverage our \datah{} benchmark to understand and evaluate cross-lingual mathematical reasoning abilities of LVLMs. Our experiments reveal a consistent performance drop when these LVLMs are evaluated on {\datah{}} (As illustrated in Figure~\ref{fig:Spider_chart}), highlighting significant shortcomings in structure-aware, cross-lingual, and intensive numerical reasoning. Although advanced paradigms like Chain-of-Thought (CoT) prompting improve performance over naive variant, they still fall short compared to their performance on English scorecards.

In summary, our contributions are three-fold: (i) We introduce \textit{\data{}}, a novel structure-aware text-centric VQA benchmark to cover for the shortcomings of existing OCR and table-based VQA benchmarks by incorporating cross-lingual, multi-image, structure-aware, and numerically rich reasoning tasks grounded in the domain of cricket analytics. (ii) We comprehensively benchmark a range of leading LVLMs (open and closed-source) across different model sizes and show that they struggle on this benchmark, revealing key limitations in structure-aware visual understanding, numerical reasoning, and cross-lingual robustness. (iii) We conduct extensive ablations incorporating specialized components such as Optical Character Recognition (OCR), Table Structure Recognition (TSR), and advanced prompting strategies including Chain-of-Thought (CoT) reasoning. While these methods improve performance, they still fall short compared to the model's strong results on conventional text-centric benchmarks, highlighting the unique difficulty of our task.

\begin{table*}[!t]
\centering
\resizebox{\textwidth}{!}
{%
\begin{tabular}{l c c c c c c}
\hline
& \multicolumn{4}{c}{Reasoning type} & & \\
\cmidrule{2-5}
\textbf{Benchmark} & \textbf{Cross-lingual} & \textbf{Multi-Image} & \textbf{Tabular} & \textbf{Mathematical} & \textbf{QA Lang.} & \textbf{VT Lang.} \\ 
\hline
\textbf{Text-centric VQA} & & & & & &  \\ 

~~TextVQA & \xmark & \xmark &  \xmark & \xmark & English & English \\
~~ST-VQA & \xmark & \xmark &  \xmark & \xmark & English & English \\
~~DocVQA  & \xmark & \xmark &  \cmark & \xmark &  English & English\\
~~EST-VQA & \xmark & \xmark &  \xmark & \xmark & English, Chinese & English, Chinese\\
~~VisualMRC & \xmark & \xmark & \cmark & \xmark & English & English\\
~~MTVQA & \xmark & \xmark &  Not a major focus & \xmark & 9 langauges & 10 languages\\
~~OCRBench & \xmark  &  \xmark &  Not a major focus & \xmark &  English & English \\
\hdashline
\textbf{Tabular VQA} & & & & & &  \\ 
~~TableVQA-Bench & \xmark&\xmark & \cmark& \cmark & English& English\\
~~TabComp &  \xmark & \xmark & \cmark & Not a major focus & English & English \\
~~ComTQA &\xmark & \xmark & \cmark & Not a major focus & English & English \\
\hdashline
\textbf{Ours} & & & & & &  \\ 

~~\textbf{\datae{}} & \cmark & \cmark & \cmark & \cmark & English  & English \\
~~\textbf{\datah{}} & \cmark & \cmark & \cmark & \cmark & English  & Hindi \\ 
~~\textbf{\data{}} & \cmark & \cmark & \cmark & \cmark & English  & English, Hindi \\ 

\hline
\end{tabular}%
}
\caption{Summary of Text-centric and Tabular VQA benchmarks, highlighting reasoning types and language support.}
\label{tab:statistics2}
\end{table*}

\section{\data{} Dataset}
We introduce \data{}, a novel dataset designed to study a visual question answering (VQA) task on cricket scorecard images. Cricket scorecard images represent unstructured yet complex tabular images. VQA on such scorecards requires structural understanding, numerical data extraction, and implicit contextual reasoning across image(s). To the best of our knowledge, our work is the first principled work on studying VQA over cricket scorecard images. Specifically, we present two sub-benchmarks under \data{}: \datae{} (with English scorecards) and \datah{} (with Hindi scorecards), with English question-answer annotations. This dataset is aimed at benchmarking the capabilities of Large Vision-Language Models (LVLMs) in performing cross-lingual deep mathematical reasoning over semi-structured content. 

\begin{table*}
\centering
\resizebox{\textwidth}{!}{
\begin{tabular}{lll}
\toprule
\textbf{Category} & \textbf{Category Name} & \textbf{Example Question} \\ \midrule
\multirow{8}{*}{C1} & \multirow{8}{*}{Direct Retrieval \& Simple Inference} & Which bowler has bowled the most wides in the match? \\
 &  & Who got out for a duck in the first innings? \\
 &  & Did any bowler take a 4-fer in the match? \\
 &  & Has {[}Batsman X{]} taken more wickets than {[}Batsman Y{]}? \\
 &  & Which bowler has conceded the most extras? \\
 &  & Who has hit the maximum sixes? \\
 &  & Does {[}Batsmax X{]} hit more sixes than {[}Batsman Y{]}? \\
 &  & How many extras were bowled in the first innings? \\ \midrule
\multirow{6}{*}{C2} & \multirow{6}{*}{Basic Arithmetic Reasoning \& Conditional Logic} & What is {[}Batsman X{]} strike rate? \\
 &  & Did {[}Batsman X{]} score better in the first innings or the second innings? \\
 &  & Which batsman scored a century in the match? \\
 &  & Which bowler took a 4-fer in the match? \\
 &  & Has {[}Batsman X{]} hit more boundaries than {[}Batsman X{]}? \\
 &  & Which batsman was dismissed for a golden duck in the match? \\ \midrule
\multirow{7}{*}{C3} & \multirow{7}{*}{Multi-step Reasoning \& Quantitative Analysis} & Which batsman had the highest strike rate (minimum 10 balls faced)? \\
 &  & Which batsman had the highest boundary percentage? \\
 &  & Which bowler had the better economy rate in the first innings? \\
 &  & Which innings had the higher run rate? \\
 &  & Which batsman had a strike rate greater than 70 in the first innings? \\
 &  & Has the same fielder caught any batsman twice? \\
 &  & Has any batsman been dismissed twice by the same bowler? \\ \bottomrule
\end{tabular}
}
\caption{Category and example questions. A full table containing statistics for each one of the single-image and multi-image questions is provided in the Appendix (Table~\ref{remain_ques}).}
\label{tab:statistics3}
\end{table*}

\data{} consists of cricket scorecards sourced from various international game formats: ODI, T20, Test Match, and popular regional leagues: the Big Bash League (BBL, Australia) and the Indian Premier League (IPL, India). We provide carefully curated QA annotations to evaluate the numerical comprehension and deep mathematical reasoning abilities of LVLMs. 
Next, we explain the dataset curation pipeline.

\noindent \textbf{Data Collection and Annotation}: We begin to collect data for our benchmark by identifying publicly available datasets and repositories that contain cricket scorecard information. The initial dataset was obtained from Kaggle\footnote{\label{datalink}\url{https://www.kaggle.com/datasets/raghuvansht/cricket-scorecard-and-commentary-dataset}}, which provides detailed cricket match statistics in CSV format. This dataset includes essential match statistics such as runs, wickets, and strike rates across different cricket formats (international game formats and regional leagues). Note that the data curated from the above-mentioned source does not contain scorecard images. 

\noindent \textbf{Scorecard Image Generation}: We employed the open-source library Weasy Print\footnote{\label{weasy}\url{https://weasyprint.org/}} to convert CSV records into visually coherent scorecard tables. The generation process was inspired by design templates from various publicly accessible sports websites, ensuring diversity in fonts, styles, and table structures. We generated two distinct types of scorecard visualizations to support different VQA scenarios: (i) single-image scorecards for limited-overs formats (ODI, T20, and league matches) containing both innings in one comprehensive image, and (ii) multi-image scorecards for Test matches, where each image contains one inning, resulting in $n$ images per match where $n$ is the number of innings in the match. This dual approach allows us to evaluate both standard single-image VQA capabilities and more complex multi-image reasoning where models must synthesize information across multiple visual inputs. The multi-image format particularly challenges models to maintain contextual awareness and perform cross-referential numerical reasoning across separate visual sources. Each scorecard image contains semi-structured tabular information such as player names, runs, balls faced, boundaries, and bowling figures, visually embedded in layouts typical of real-world cricket statistics. More details regarding the specific fonts, structural variations, and template designs are provided in the Appendix ~\ref{template_design}.

\noindent \textbf{Data Translation}: To create images for \datah{} benchmark, we translate the English CSV records into Hindi using Google Translate\footnote{\url{https://translate.google.com}} and follow a similar synthetic image generation procedure as of \datae{}. 
The \datah{} sub-benchmark is introduced to evaluate the cross-lingual semantic landscape of LVLMs associated with answering complex visual questions. Note that the translation process was performed at the cell level, ensuring that cricket-specific terms and numerical patterns remained consistent across languages. Additionally, we conducted a manual review of translated records to account for any potential translation inaccuracies and domain-specific inconsistencies.

\begin{figure}[!t]
    \includegraphics[width=.48\textwidth]{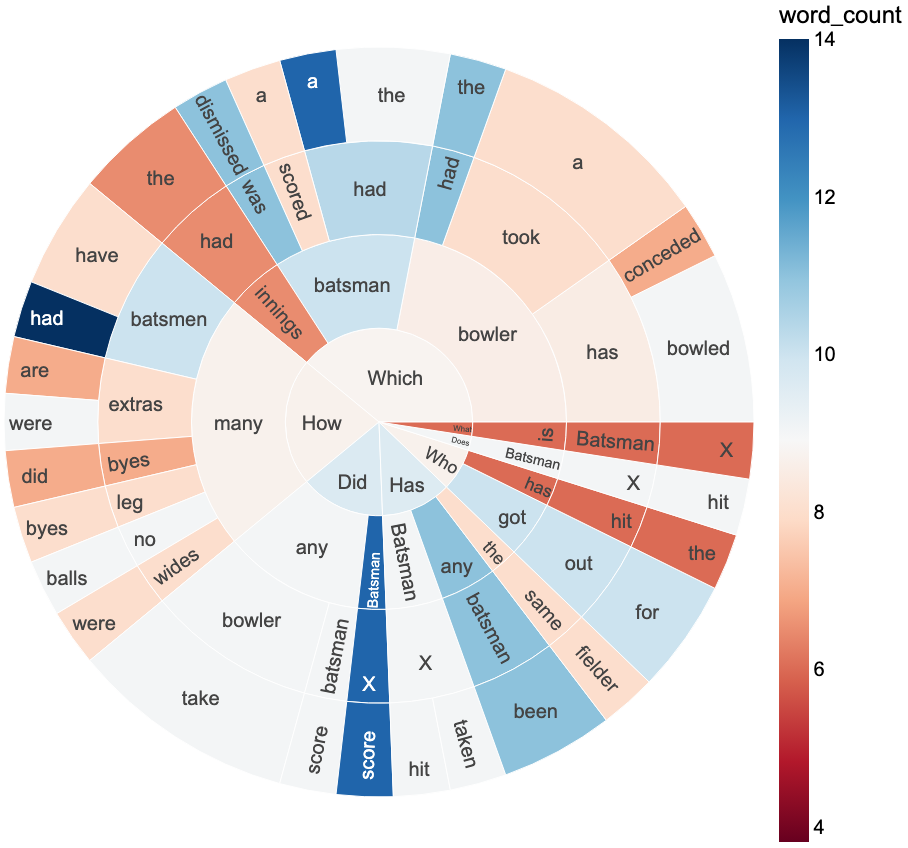}
    \caption{Distribution of questions based on their first few words in the dataset, illustrating common question prefixes after preprocessing.}
\label{fig:third_diagram}
\end{figure}

\noindent \textbf{Question-Answer Generation:}
We leverage cricket scorecard images to construct the \data{} dataset, designed to evaluate Large Vision-Language Models (LVLMs) on structure-aware, cross-lingual, and numerically intensive reasoning tasks. We manually designed question templates and categorized them into three categories depending on the complexity associated with answering these questions. The categories are: \textbf{(i) Direct Retrieval \& Simple Inference - C1:} In this task, questions target direct information extraction from the scorecard image. For example, given an image containing the entry: \textit{Aaron Finch | 89 | 52 | 4 | 7}, we generate a question: \textit{``Who hit the most sixes?''} with the answer \textit{``Aaron Finch''}. The model must read the image, interpret the tabular layout, understand the required context, and identify the relevant value. \textbf{(ii) Basic Arithmetic Reasoning \& Conditional Logic - C2:} Here, questions require numerical reasoning based on arithmetic operations or conditional checks applied to one or more rows in the image. For example, from a scorecard showing: \textit{Virat Kohli | 94 runs | 50 balls}, we generate the question: \textit{``What is Virat Kohli’s strike rate?''} with the answer \textit{``188.0''}, computed using: \(\text{Strike Rate} = \frac{\text{Runs}}{\text{Balls}} \times 100\). The model must correctly localize relevant cells, extract values, and apply the correct reasoning. \textbf{(iii) Multi-step Reasoning \& Quantitative Analysis - C3:} This task involves combining information across multiple players or sections in the scorecard. For instance, to answer: \textit{``Who has the highest boundary percentage?''}, the model needs to compute \(\frac{(4s \times 4 + 6s \times 6)}{\text{Total Runs}} \times 100\) for each player and select the maximum. This requires layout-aware text extraction, numerical computation, and multi-row comparison across the image.

Few question templates across the three categories are shown in Table~\ref{tab:statistics3}. Detailed questions and statistics under all three categories are shown in Appendix  ~\ref{remain_ques}. 

\noindent \textbf{Answer Extraction via SQL:} To ensure accuracy and consistency in answer generation, we used SQL queries to derive answers directly from the structured CSV data. This approach minimized manual errors and ensured the traceability of answers back to the original data. The SQL queries were formulated based on the question type and corresponding data structure. For instance:

\begin{itemize}
    \item To retrieve highest boundary percentage:\\ \texttt{SELECT Batsman\_Name FROM batting WHERE Innings = 1 AND Balls > 0 ORDER BY ((([4s]*4 + [6s]*6) * 100.0 / Runs)) DESC LIMIT 1;}
    \item To retrieve better economy rate in innings 1:\\ \texttt{SELECT Bowler\_Name, ROUND((SUM(Runs) * 1.0 / SUM(Over)), 2) AS Economy\_Rate FROM bowling WHERE Innings = 1 GROUP BY Bowler\_Name ORDER BY Economy\_Rate ASC LIMIT 1;}
\end{itemize}
SQL queries for every question template in \data{} are shown in Table ~\ref{Question:SQL} in the Appendix. Further, the question-answer pairs are subjected to manual verification for possible factual and mathematical errors. 

Further, we categorized answers into four categories, namely, (i) Binary (Yes/No), (ii) Numerical, (iii) Categorical (1/2/3/4 for innings-based questions), and (iv) Open-ended (Person names). Detailed statistics of \data{} for are shown in the Figure~\ref{fig:datasetAnalysis} (a). Further, a selection of a few QA samples for each of the answer categories is shown in Table~\ref{Answerwise} in the Appendix.

\section{Experiments}

\begin{table*}[!t]
    \centering
    \resizebox{0.8\textwidth}{!}{
    \begin{tabular}{l c c c c c c c c c c}
        \toprule
        & \multicolumn{4}{c}{MMCricBench-E-1.5K} 
        & \multicolumn{4}{c}{MMCricBench-H-1.5K} 
        & 
        &  \\
        \cmidrule(lr){2-5} \cmidrule(lr){6-9}
        \textbf{Model [\#params]} 
        & C1 & C2 & C3 & Avg. 
        & C1 & C2 & C3 & Avg. 
        & \(\Sigma\uparrow\) 
        & \(\Delta\downarrow\) \\
        \midrule
        \multicolumn{11}{c}{\cellcolor[gray]{0.9}\textbf{Open-source}} \\
        \midrule
        \textit{Small VLM ($\leq$ 3B params)} &  &  &  &  &  &  &  &  &  &  \\
        ~~SmolVLM [500M]   & 19.5 & 21.6 & 15.9 & 19.2 
                           & 20.4 & 12.9 & 24.3 & 19.0 
                           & 19.1 
                           & 0.2 \\
        ~~Qwen2.5VL [3B]   & 38.7 & 40.1 & 41.7 & 40.2 
                           & 39.8 & 24.5 & \textbf{35.5} & 33.3 
                           & 36.8 
                           & 6.9 \\
        \hdashline
        \textit{Large VLM (params >3B and <10B)}        &  &  &  &  &  &  &  &  &  &  \\
        ~~LLaVA-NeXT [7B]  & 40.2 & 10.8 & \textbf{33.9} & 28.3 
                           & 35.7 & 10.8 & 33.3 & 26.6 
                           & 27.4 
                           & 1.7 \\
        ~~mPlugDocowl2 [8B]& 33.9 & 13.9 & 14.2 & 20.7 
                           & 33.6 & 13.7 & 12.3 & 19.9 
                           & 20.3 
                           & 0.8 \\
        ~~Qwen2.5VL [7B]   & \textbf{64.6} & \textbf{52.1} & 30.6 & \textbf{49.1} 
                           & \textbf{62.7} & \textbf{39.8} & 25.2 & \textbf{42.6} 
                           & \textbf{45.8} 
                           & 6.5 \\
        ~~InternVL-2 [8B]  & 33.6 & 26.3 & 28.2 & 29.4 
                           & 28.5 & 16.4 & 25.2 & 23.4 
                           & 26.4
                           & 6.0 \\
        \hdashline
        \textit{X-Large VLM (>10B)}     &  &  &  &  &  &  &  &  &  &  \\
        ~~Llama-3.2-V [11B]& 26.7 & 35.3 & 19.8 & 27.3 
                           & 25.2 & 26.9 & 22.2 & 24.8 
                           & 26.0 
                           & 2.5 \\
        \midrule
        \multicolumn{11}{c}{\cellcolor[gray]{0.9}\textbf{Closed-source}} \\
        \midrule
        GPT-4o  & 56.0 & 65.1 & 50.6 & 57.3 & 54.6 & 49.7 & 30.9 & 45.1 & 50.5 & 12.2 \\
        \bottomrule
    \end{tabular}
    }
    \caption{\label{tab:singleimage_res} Results on single-image questions split of \data{}.}
\end{table*}

\begin{table*}[!t]
    \centering
    \resizebox{0.8\textwidth}{!}{
    \begin{tabular}{l c c c c c c c c c c}
        \toprule
        & \multicolumn{4}{c}{MMCricBench-E-1.5K} 
        & \multicolumn{4}{c}{MMCricBench-H-1.5K} 
        & 
        &  \\
        \cmidrule(lr){2-5} \cmidrule(lr){6-9}
        \textbf{Method [\#params]} 
        & C1 & C2 & C3 & Avg. 
        & C1 & C2 & C3 & Avg. 
        & \(\Sigma\uparrow\) 
        & \(\Delta\downarrow\) \\
        \midrule
        \textbf{LLMs$+$OCR}            &      &      &      &      &      &      &      &      &      &      \\
        ~~Llama-3.2 [3B] & 32.1 & \textbf{31.4} & \textbf{22.8} & \textbf{28.8} & 24.1 &  7.4 & \textbf{18.3} & 16.6 & 22.7 & 12.2 \\
        ~~Qwen2.5 [3B] & \textbf{36.6} & \textbf{31.4} & 16.5 & 28.2 & \textbf{34.2} &  \textbf{13.1} & 13.5 & \textbf{20.3} & \textbf{24.2} & 7.9 \\
        \hdashline
        \textbf{VLMs Chain-of-Thought} &      &      &      &      &      &      &      &      &      &      \\
        ~~Qwen2.5VL [7B]  & 69.1 & 55.7 & 36.0 & 53.6 & 65.2 & 40.7 & 31.5 & 45.8 & 49.7 &  7.8 \\
        \bottomrule
    \end{tabular}
    }
    \caption{\label{tab:ocr_added_result}Results on single-image of our ablation: LLMs+OCR vs VLMs on \data{}.}
\end{table*}

\label{sec:exps}

\noindent \textbf{Baselines.} We chose the VLMs from three selection criteria: \textbf{(a) VLMs with no OCR-aware tasks} during their pretraining or instruction tuning stages: LLaVA-Next~\cite{liu2024llava}, and \textbf{(b) VLMs based on the size of their parameters}: \textit{(i) Small VLMs (SVLMs)} with parameters less than 5B: SmolVLM-500M~\cite{marafioti2025smolvlm}, Qwen2.5VL-2B~\cite{wang2024qwen2}, \textit{(ii) Large VLMs (LVLMs)} with parameters between 5B-14B: InternVL2-8B~\cite{chen2024internvl}, Qwen2.5VL-7B~\cite{bai2025qwen2}, mPLUG-DocOwl2~\cite{hu2024mplugdocowl2}, \textit{(c) X-Large VLMs} with parameters greater than 14B: Llama-3.2-V-11B~\cite{grattafiori2024llama} and (iii) \textbf{closed-source VLMs}: GPT-4o~\cite{openai2024gpt4api}. 

\subsection{Result and Discussion}
\noindent \textbf{Performance of Open-Source Models Across Scales}:
Tables~\ref{tab:singleimage_res} and~\ref{tab:multiimage_res} present results for single-image and multi-image setups, revealing a consistent trend: model scale has a notable impact on performance across all question categories. Larger models generally outperform their smaller counterparts, with more pronounced gains on complex reasoning categories such as C2 (arithmetic) and C3 (multi-hop reasoning). For instance, Qwen2.5VL-7B~\cite{bai2025qwen2} significantly outperforms its smaller 3B variant across all settings, with an average performance gap of 8.5 points. While this scaling advantage is particularly evident in higher-complexity tasks, the gains are less pronounced on simpler C1 (retrieval-based) questions, as expected.

\begin{table*}[!t]
    \centering
    \resizebox{0.8\textwidth}{!}{
    \begin{tabular}{l c c c c c c c c c c}
        \toprule
        & \multicolumn{4}{c}{MMCricBench-E-1.5K} 
        & \multicolumn{4}{c}{MMCricBench-H-1.5K} 
        &  
        &  \\
        \cmidrule(lr){2-5} \cmidrule(lr){6-9}
        \textbf{Model [\#params]} 
        & C1 & C2 & C3 & Avg. 
        & C1 & C2 & C3 & Avg. 
        & \(\Sigma\uparrow\) 
        & \(\Delta\downarrow\) \\
        \midrule
        \multicolumn{11}{c}{\cellcolor[gray]{0.9}{\textbf{Open-source}}}\\
        \midrule
        \textit{Small VLM ($\leq$ 3B params)} 
          &      &      &      &      &      &      &      &      &      &      \\
        ~~SmolVLM [500M] 
          & 14.4 & 10.8 & 10.2 & 11.8 
          & 20.0 &  6.0 &  9.0 & 11.6 
          & 11.7 
          &  0.2 \\
        ~~Qwen2.5VL [3B] 
          & 34.1 & 35.3 & 24.1 & 31.2 
          & 27.5 & 19.8 & 18.7 & 22.0 
          & 26.6 
          &  9.2 \\
        \hdashline
        \textit{Large VLM (params >3B and <10B)} 
          &      &      &      &      &      &      &      &      &      &      \\
        ~~LLaVA-NeXT [7B] 
          & 27.5 &  6.6 & 14.4 & 16.2 
          & 24.5 &  5.4 & 14.5 & 14.8 
          & 15.5 
          &  1.4 \\
        ~~mPlugDocowl2 [8B] 
          & 24.7 &  7.5 & 13.2 & 15.2 
          & 23.3 &  7.1 & 12.6 & 14.4 
          & 14.8 
          &  0.8 \\
        ~~Qwen2.5VL [7B] 
          & \textbf{41.9} & \textbf{41.9} & 27.1 & \textbf{37.0} 
          & \textbf{37.7} & \textbf{33.5} & \textbf{25.3} & \textbf{32.2} 
          & \textbf{34.6} 
          &  4.8 \\
        ~~InternVL-2 [8B] 
          & 29.3 & 5.4 & 21.1 & 18.6
          & 28.1 & 4.8  & 21.7 & 18.2
          &  18.4
          &  0.4 \\
        \hdashline
        \textit{X-Large VLM (>10B)} 
          &      &      &      &      &      &      &      &      &      &      \\
        ~~Llama-3.2-V [11B] 
          & 34.7 & 14.3 & \textbf{29.5} & 26.2 
          & 29.3 & 11.3 & 20.4 & 20.4 
          & 23.3 
          &  5.8 \\
        \midrule
        \multicolumn{11}{c}{\cellcolor[gray]{0.9}{\textbf{Closed-source}}}\\
        \midrule
        GPT-4o 
          & 50.3 & 61.1 & 40.4 & 50.6 
          & 39.5 & 53.8 & 37.3 & 43.6 
          & 47.1 
          &  7.0 \\
        \bottomrule
    \end{tabular}
    }
    \caption{\label{tab:multiimage_res} Results on multi-image questions split of \data{}.}
\end{table*}

\begin{table*}[!t]
    \centering
    \resizebox{0.8\textwidth}{!}{
    \begin{tabular}{l c c c c c c c c c c}
        \toprule
        & \multicolumn{4}{c}{MMCricBench-E-1.5K} 
        & \multicolumn{4}{c}{MMCricBench-H-1.5K} 
        &  
        &  \\
        \cmidrule(lr){2-5} \cmidrule(lr){6-9}
        \textbf{Method [\#params]} 
        & C1 & C2 & C3 & Avg. 
        & C1 & C2 & C3 & Avg. 
        & \(\Sigma\uparrow\) 
        & \(\Delta\downarrow\) \\
        \midrule
        \textbf{LLMs$+$OCR}            &      &      &      &      &      &      &      &      &      &      \\
        ~~Llama-3.2 [3B] & 24.5 & 17.9 & 25.9 & 22.8 & 18.5 &  1.8 & 14.4 & 11.6 & 17.2 & 11.2 \\
        ~~Qwen2.5 [3B]  & \textbf{30.5} & \textbf{24.5} & \textbf{27.7} & \textbf{27.6} & \textbf{23.3} &  \textbf{10.7} & \textbf{22.8} & \textbf{19.0} & \textbf{23.3} & 8.6 \\
        \hdashline
        \textbf{VLMs Chain-of-Thought} &      &      &      &      &      &      &      &      &      &      \\
        ~~Qwen2.5VL [7B] & 40.7 & 40.7 & 22.9 & 34.8 & 36.5 & 29.9 & 23.5 & 30.0 & 32.4 &  4.8 \\
        \bottomrule
    \end{tabular}
    }
    \caption{\label{tab:ocr_added_multi_result}Results on multi-image of our ablation: LLMs+OCR vs VLMs on \data{}.}
\end{table*}

\noindent \textbf{Closed-Source vs Open-Source Models}:
Closed-source models, notably GPT-4o, consistently outperform open-source models across both the English (\datae{}) and Hindi (\datah{}) subsets. On single-image questions, GPT-4o achieves the highest average accuracy of 57.3\% on English and 45.1\% on Hindi, while in the multi-image setting, it scores 50.6\% on English and 43.6\% on Hindi. This reflects a clear cross-lingual drop of 12.2 and 7.0 points in the single- and multi-image settings, respectively. Although GPT-4o is not immune to the challenges posed by script variation, it still outperforms the closest open-source model Qwen2.5VL-7B by an average margin of 8.2 points across all tasks and subsets. These results highlight the robustness gap that remains between open and closed-source models, particularly in structured, cross-lingual VQA settings.

\noindent \textbf{Comparison of cross-lingual capabilities}: Models consistently exhibit a significant performance drop when transitioning from English to Hindi scorecards, particularly in categories requiring arithmetic reasoning (C2) and multi-step reasoning (C3). This decline highlights the limitations of cross-lingual generalization that scaling alone fails to address. For instance, GPT-4o, the strongest overall performer shows a substantial drop of 12.2 points (single-image) and 7.0 points (multi-image) on average when evaluated on Hindi scorecards. Similarly, Qwen2.5VL-7B experiences a 6.5 to 6.9 point decrease across both subsets. These degradations indicate that even state-of-the-art models with strong English capabilities are not robust to script variation in visually embedded text. Our findings suggest that effective VQA on cricket scorecards requires a combination of table structure understanding, OCR, and visual text grounding—capabilities that current models struggle to achieve in non-Latin scripts and low-resource visual text languages like Hindi.

\subsubsection{Ablations}

\noindent \textbf{LLMs + OCR}: To isolate the role of visual perception in scorecard-based VQA, we evaluate a baseline that combines OCR with text-only large language models (LLMs). Specifically, we extract text from scorecard images using the Tesseract OCR engine~\cite{smith2007overview} and feed the output into two LLMs: LLaMA-3.2-3B~\cite{dubey2024llama} and Qwen2.5-3B~\cite{yang2024qwen2technicalreport}. This setting evaluates whether textual cues alone are sufficient to reason over cricket scorecards. As shown in Tables~\ref{tab:ocr_added_result} and~\ref{tab:ocr_added_multi_result}, both models perform significantly worse than vision-language models. On average across \data{}, LLaMA-3.2-3B exhibits a performance drop of 4.7\%, while Qwen2.5-3B shows a much larger drop of 16.5\% compared to their vision counterparts. These results highlight the limitations of OCR+LLM pipelines: despite having access to textual input, these models struggle to capture structural cues such as column alignment and row grouping that are essential for tabular reasoning. The English-Hindi gap remains wide, showing that OCR-based pipelines struggle in cross-lingual, visually complex settings.

\noindent \textbf{CoT Prompting vs. Regular Prompting}:
Applying Chain-of-Thought (CoT) prompting to Qwen2.5VL-7B improves overall performance in the single-image setting, with accuracy increasing from 45.8\% to 49.7\%. This gain is especially notable in reasoning-heavy categories such as arithmetic (C2) and multi-step (C3), indicating that CoT helps the model decompose complex queries into interpretable steps. However, in the multi-image setting, overall performance drops slightly from 34.6\% to 32.4\%, suggesting that CoT may not transfer well when reasoning must span multiple visual contexts. While CoT improves reasoning behaviour, the cross-lingual gap still remains.

\section{Comparison with Related Work}
\noindent \textbf{LVLMs for VQA over text images}: Recent advancements of large vision-language models (LVLMs) have transformed visual question-answering (VQA) tasks into gaining impressive zero-shot performance across diverse scenarios \cite{openai2024gpt4api,yang2024qwen2technicalreport,chen2024internvl,liu2024llava} including text-centric VQA. On these lines, DocPedia \cite{feng2024docpedia} processes high-resolution inputs without increasing token sequence length.  mPLUG-DocOwl\cite{ye2024mplug}, Qwen2-VL \cite{wang2024qwen2}, and TextMonkey \cite{liu2024textmonkey} further leverage publicly available document VQA datasets to boost text performance.\ Extensions of the LLaVA \cite{liu2024visual} framework such as LLaVAR \cite{zhang2023llavar}, InternVL \cite{chen2024internvl}, and UniDoc \cite{feng2023unidoc} have broadened LVLM capabilities into document understanding by predicting both textual content and spatial coordinates, thereby setting a new benchmark for text-centric VQA. Despite these significant strides, LVLMs fall short in complex tasks like \data{} as discussed in Section~\ref{sec:exps}.

\noindent \textbf{Text-centric VQA}:
The existing text VQA datasets TextVQA \cite{singh2019towards}, ST-VQA \cite{biten2019scene}, DocVQA \cite{mathew2021docvqa}, and VisualMRC \cite{tanaka2021visualmrc} solely focus on the English language. While EST-VQA \cite{wang2020general} and MTVQA \cite{tang2024mtvqa} are multilingual, they do not cover low-resource visual languages e.g. Hindi. Further, existing datasets either primarily focus on single-image QA or lack questions that require structure-aware mathematical reasoning (summarized in Table \ref{tab:statistics2}). We aim to address this gap.

\noindent \textbf{Models and Datasets for Table VQA}:
While benchmark datasets like TableVQA-Bench ~\cite{kim2024tablevqa}, TabComp ~\cite{gautam2025tabcomp}, and ComTQA ~\cite{zhao2024tabpedia} exist for VQA over table images, they are all English-focused with answers directly in the images. However, table image datasets to evaluate the cross-lingual mathematical reasoning capabilities of LVLMs remain underexplored.

\noindent \textbf{Multi-image VQA:}  
Several benchmarks~\cite{talmor2021multimodalqa,mathew2021docvqa,imageset_vqa,webQA21,penamakuri2023answer,visualhaystacks} explore reasoning across multiple image. However, these tasks largely overlook structure-aware tabular understanding, numerical reasoning, and cross-lingual robustness, which are central to our setting. In contrast, we include a dedicated multi-image subset within \data{}, where answering a question requires aggregating statistics across multiple images representing different innings of a match, thereby combining tabular, numerical, and cross-lingual reasoning.

\noindent \textbf{Table Reasoning Ability of LLMs}:
LLMs and multimodal LLMs (MLLMs) are evaluated in \cite{deng2024tables} using tables presented as either text or images, finding that text-based representations yield better results, while image-based table reasoning remains weak for current models.
To enhance reasoning, \cite{lu2024tart} introduced TART, a tool-augmented framework that enables step-by-step table question answering by integrating LLMs with symbolic tools. Similarly, \cite{nahid2024normtab} proposed TabSQLify, which improves efficiency by decomposing large tables into smaller, relevant segments using text-to-SQL conversion.
Furthermore, \cite{zhao2022reastap} proposed ReasTAP, a pretraining strategy using synthetic table reasoning examples to inject structured reasoning ability into LLMs.
Despite these advances, most research focuses on structured tables in English. A critical gap remains in (i) table reasoning in low-resource languages (such as Hindi in our dataset), particularly visually complex, domain-specific formats like cricket scorecards, (ii) evaluating multi-step reasoning and conditional logic in understanding the tabular content. Our work addresses this need by introducing \datah{}, a benchmark designed to push the limits of visual-text reasoning in Hindi.

\section{Conclusion}
We presented \data{}, a novel benchmark for VQA on cricket scorecards that addresses critical gaps in existing datasets by incorporating cross-lingual understanding, multi-image reasoning, and domain-specific numerical analysis. Our evaluation across \datae{} (English) and \datah{} (Hindi) scorecards reveals a significant performance disparity among state-of-the-art LVLMs. While these models show reasonable proficiency with English scorecards, they struggle substantially with Hindi variants despite identical information content. Even advanced prompting strategies like CoT fail to bridge this performance gap. These findings highlight a critical weakness in cross-lingual visual reasoning capabilities, highlighting the need for more robust models that can effectively process structured numerical data across language boundaries. As AI applications expand globally, addressing these limitations becomes increasingly crucial. \data{} provides researchers with a challenging testbed for advancing LVLM capabilities beyond English-centric contexts, particularly in domains requiring precise analysis of semi-structured information.

\section{Limitations}
Despite the strengths of \data{} in evaluating structure-aware and cross-lingual visual question answering, several limitations persist. First, the dataset's linguistic scope is limited to English and Hindi, leaving out other regional scripts and languages prevalent in cricket contexts. Second, the use of synthetically generated scorecards, while visually coherent, may not fully capture the complexity and noise present in real-world documents.

\section*{Ethical Considerations}

Our benchmark, \data{}, is synthetically generated using publicly available cricket statistics, with no private or sensitive personal information involved. All scorecard data is derived from open datasets (e.g., Kaggle) and only includes publicly known player names and match events. We translate content using automated tools (e.g., Google Translate), and manually verify for correctness to minimize cultural or linguistic bias.

While our dataset uses Hindi as a representative low-resource script for cross-lingual evaluation, we acknowledge the limitations of focusing only on English-Hindi and encourage future extensions to other regional languages and scripts. Additionally, though we simulate realistic scorecards, real-world images may include noise, varied layouts, or OCR artifacts that are not fully captured in our synthetic setup. Our work aims to support fair and inclusive evaluation of vision-language models in global contexts. No human annotators were subjected to sensitive or harmful content during data creation, and no demographic or identity information is used or inferred in this study.

\section*{Acknowledgments}
Abhirama is deeply grateful to his PhD advisor, Dr. Anand Mishra, for his unwavering guidance and mentorship throughout his PhD journey, which has profoundly shaped his research outlook and provided the foundation that enabled this work. Abhirama is supported by the Prime Minister’s Research Fellowship (PMRF), Ministry of Education, Government of India.

\bibliography{custom}

\appendix

\section{Appendix}

\subsection{Implementation Details}
We conduct all of our experiments on the baseline VLMs in a zero-shot setting, with their default setting provided in their respective implementations. When prompted with these methods, we faced two challenges: (i) verbose answers and (ii) digits written in text, e.g. Fifth in place of 5. To overcome these challenges and generate precise and concise answers, we added a brief instruction to the prompt: `Answer precisely in 1-2 words, answer in digits when required' before the main question. We conducted all our experiments on a cloud machine with 3 A6000 Nvidia GPUs (48 GB each) rented from online cloud GPU provider TensorDock~\cite{tensordock2024}.

\subsection{Cricket Specific Terms and Their Calculations}
In cricket, performance metrics help quantify a player's efficiency in both batting and bowling. Two key metrics are the Strike Rate and the Economy Rate. The following explanations and formulas provide a detailed understanding of these terms.

\subsubsection{Strike Rate}
The Strike Rate is primarily used to measure a batsman's scoring efficiency. It represents the average number of runs scored per 100 balls faced, indicating how quickly a batsman can accumulate runs.\\

\textbf{Calculation:}
The basic formula for Strike Rate is:
\[
\text{Strike Rate (SR)} =  \frac{\text{Total Runs Scored}}{\text{Total Balls Faced}}  \times 100
\]

\textbf{Example:}
For instance, if a batsman scores 50 runs from 40 balls, the Strike Rate is calculated as:
\[
\text{SR} = \frac{50}{40}  \times 100 = 125
\]
This means that, on average, the batsman scores 125 runs for every 100 balls faced. A higher strike rate reflects a more aggressive and effective scoring approach.

\subsubsection{Economy Rate}
The Economy Rate measures a bowler’s efficiency by calculating the average number of runs conceded per over. An over in cricket typically consists of 6 legal deliveries.

\textbf{Calculation: }
The basic formula for Economy Rate is:
\[
\text{Economy Rate (Econ)} = \frac{\text{Total Runs Conceded}}{\text{Total Overs Bowled}}
\]
If the data is provided in terms of balls bowled rather than overs, the formula is adjusted by converting balls to overs:
\[
\text{Economy Rate (Econ)} = \frac{\text{Total Runs Conceded}}{\text{Total Balls Bowled}} \times 6
\]

\textbf{Example:}
Consider a bowler who concedes 30 runs in 10 overs. The Economy Rate is:
\[
\text{Econ} = \frac{30}{10} = 3.0 \text{ runs per over}
\]
A lower economy rate suggests that the bowler is effective at limiting the opposing team's scoring.

\subsubsection{Summary}
Understanding these calculations is fundamental for analyzing cricket performance:
\begin{itemize}
    \item The Strike Rate provides insight into a batsman's ability to score quickly, which is especially valuable in limited-overs formats.
    \item The Economy Rate evaluates a bowler's performance by highlighting how few runs they allow per over, thus reflecting their effectiveness in containing the opposition's scoring.
\end{itemize}

These metrics are essential for comparing player performances across different matches and cricket formats, offering a standardized way to assess and discuss efficiency in both batting and bowling.

\subsubsection{Rationale for Multi-Images}
Cricket matches played over multiple innings often contain statistics that span beyond a single table or image. For instance, Test matches commonly have four innings across five days, with runs, wickets, and partnerships distributed across these innings. A single static image may not encapsulate the full statistical narrative, necessitating a shift toward a multi-image structure. LVLMs must then establish logical connections across these images to accurately answer questions involving cumulative statistics or cross-inning performance comparisons.

\begin{figure}[!hb]
\centering
    \subfloat[C1]{\includegraphics[width=0.8\linewidth]{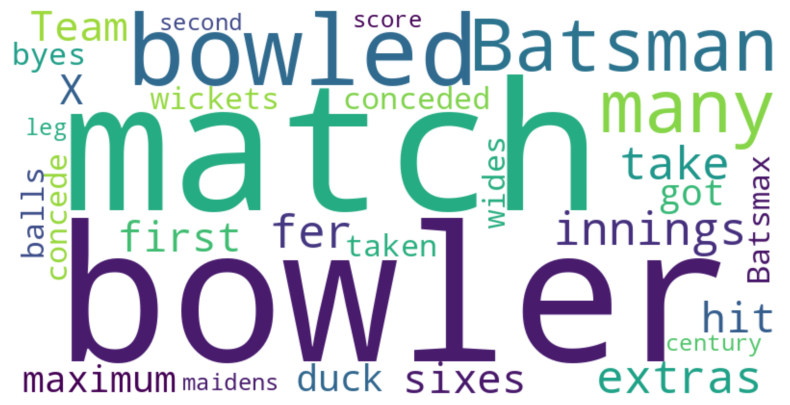}}\\
    
    \subfloat[C2]{\includegraphics[width=0.8\linewidth]{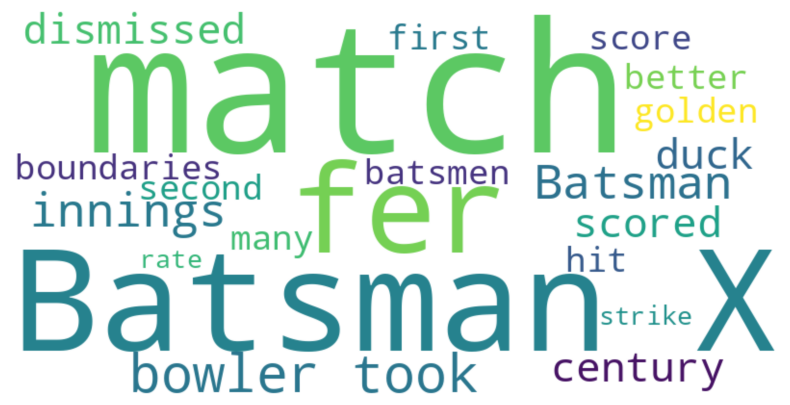}}\\
    
    \subfloat[C3]{\includegraphics[width=0.8\linewidth]{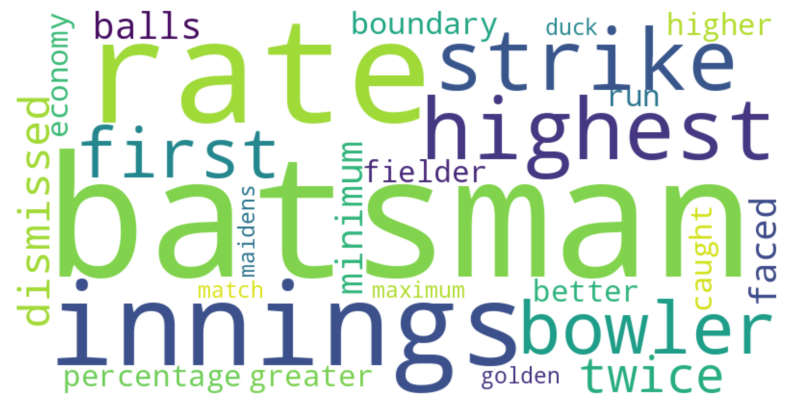}}\\
    \caption{Word cloud of category-wise questions.}
    \label{fig:images}
\end{figure}

\begin{table}
\centering
\begin{tabular}{|l|}
\hline
\textbf{Country}      \\ \hline
Afghanistan  \\ \hline
Australia    \\ \hline
Bangladesh   \\ \hline
England      \\ \hline
India        \\ \hline
Ireland      \\ \hline
New Zealand  \\ \hline
Pakistan     \\ \hline
South Africa \\ \hline
Sri Lanka    \\ \hline
West Indies  \\ \hline
Zimbabwe     \\ \hline
Netherlands  \\ \hline
\end{tabular}
\caption{Distribution of Scorecards Across 13 Countries.}
\label{Country}
\end{table}

\subsection{Scorecard Image Template Design}
\label{template_design}
The HTML/CSS template used to render each match's batting and bowling scorecards is defined via a Jinja2 template and styled to ensure consistent layout and visual separation of sections. Images are generated by rendering this HTML to PDF via WeasyPrint, converting single‐page PDFs to 300 DPI PNGs, and cropping whitespace. Below, we list all key design parameters in Table ~\ref{tab:template}.
\begin{table*}[!t]
  \centering
  \small
  \begin{tabular}{@{} l l p{0.65\textwidth} @{}}
    \toprule
    \textbf{Category} 
      & \textbf{Ingredient} 
      & \textbf{Specification} \\
    \midrule
    \multirow{3}{*}{Setup}
      & Page setup \& font
      & 20\,px margins; white background; Arial, sans-serif font. \\
      & Table layout
      & Full-width tables; collapsed borders; centered; 20\,px vertical margins. \\
      & Cell padding
      & 1\,px padding on all header and data cells. \\
    \midrule
    \multirow{6}{*}{Styling}
      & Header row styling
      & Custom background color; black text; 14\,px font; left-aligned. \\
      & Data cell styling
      & 14\,px font; centered text. \\
      & Column widths
      & First column left-aligned (min-width 120 px); others centered (min-width 60 px). \\
      & Team-name banner
      & Bold 12 px text on customizable background; 5 px padding and vertical margins. \\
      & Section separation
      & 1 px bottom border + extra spacing between innings. \\
      & Special rows
      & Bold white rows for “Extras” and “Total.” \\
    \midrule
    \multirow{4}{*}{Color variants}
      & Variant 1
      & Banner \#DA8EE7; header \#CCCCFF. \\
      & Variant 2
      & Banner \#E8CCFF; header \#CCE7FF. \\
      & Variant 3
      & Banner \#D0CCFF; header \#E8CCFF. \\
      & Variant 4
      & Banner \#CCFFE7; header \#CCFFCC. \\
    \bottomrule
  \end{tabular}
  \caption{Template ingredients for scorecard image generation.}
  \label{tab:template}
\end{table*}

\subsubsection{Country Diversity}
Cricket is a global sport played across continents, with scorecards reflecting diverse naming conventions, team compositions, and performance statistics. Incorporating scorecards from 13 countries ensures that the dataset captures these variations, providing a comprehensive representation of cricketing data across different regions and formats. The country list is in table \ref{Country}.

\begin{figure}
    \includegraphics[width=.48\textwidth]{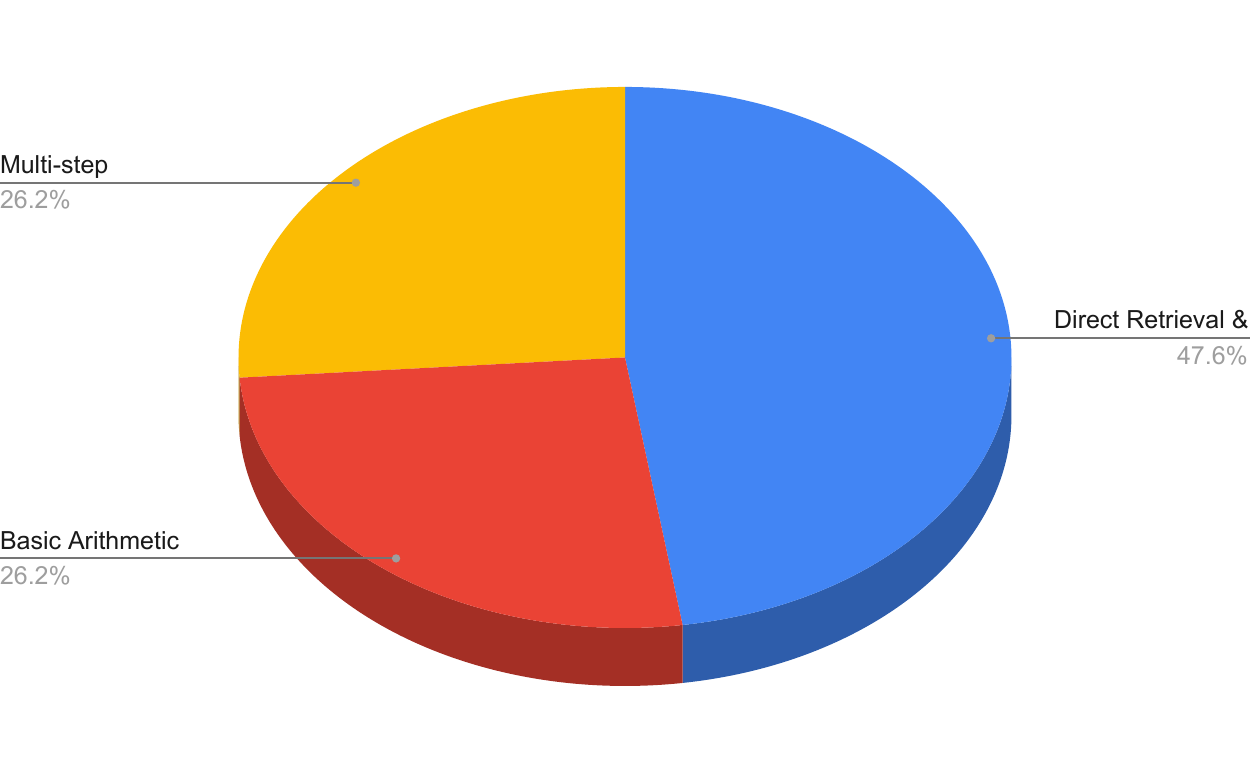}
    \caption{Category Distribution for Batting and Bowling Questions.}
\label{fig:fourth_diagram}
\end{figure}

\begin{figure}
  \scriptsize
   \centering
   \includegraphics[width=0.48\textwidth]{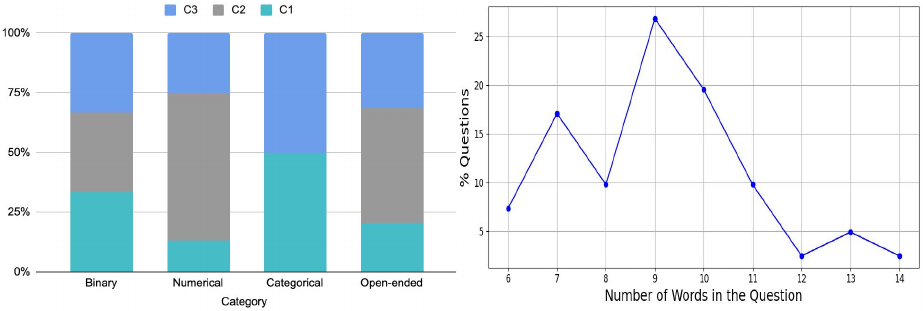}
     \caption{\data{} questions and answers analysis: (a) Answer distribution over various question categories, (b) Distribution of the number of words across questions. 
     }
     \label{fig:datasetAnalysis}
\end{figure}

\begin{table*}[t]
\centering
\resizebox{\textwidth}{!}{
\begin{tabular}{cccccccccccccccccc}
\hline
Model & \multicolumn{9}{c}{\datae{}} & \multicolumn{8}{c}{\datah{}} \\ \cmidrule(lr){3-10} \cmidrule(lr){11-18}
 & \multicolumn{5}{c}{Single} & \multicolumn{4}{c}{Multi} & \multicolumn{4}{c}{Single} & \multicolumn{4}{c}{Multi} \\ \cmidrule(lr){3-6} \cmidrule(lr){7-10} \cmidrule(lr){11-14} \cmidrule(lr){15-18}
 & Cat. & i & ii & iii & iv & i & ii & iii & iv & i & ii & iii & iv & i & ii & iii & iv \\ \hline
\multirow{3}{*}{GPT4o} & C1 & 84.1 & 30.6 & \textbf{50.0} & 45.0 & \textbf{79.6} & 28.33 & 16.6 & 50.0 & 81.3 & 15.9 & \textbf{50.0} & \textbf{30.4} & 71.1 & 15.0 & \textbf{41.6} & 27.7 \\
 & C2 & 67.9 & \textbf{94.4} & NA & \textbf{50.2} & 60.0 & \textbf{75.00} & NA & \textbf{56.90} & 55.5 & \textbf{75.0} & NA & 27.1 & 33.3 & \textbf{80.56} & NA & \textbf{48.2} \\
 & C3 & 57.6 & 65.8 & 44.7 & 35.7 & 69.23 & 17.14 & NA & 31.6 & 35.7 & 33.3 & 36.8 & 16.3 & 67.3 & 25.7 & NA & 22.78 \\ \hline
\multirow{3}{*}{Qwen2.5VL {[}7B{]}} & C1 & \textbf{87.3} & 68.1 & 0.0 & 25.6 & 72.8 & 21.6 & \textbf{33.3} & 27.7 & \textbf{85.8} & 69.9 & 0.0 & 18.2 & \textbf{76.2} & 20.0 & 16.6 & 11.1 \\
 & C2 & 55.5 & 69.5 & NA & 42.5 & 46.6 & 69.4 & NA & 33.3 & 32.7 & 72.8 & 0.0 & 19.1 & 46.6 & 63.8 & NA & 22.4 \\
 & C3 & 58.9 & 10.7 & 5.2 & 30.1 & 59.6 & 2.8 & Na & 16.4 & 51.5 & 10.7 & 13.1 & 18.1 & 67.3 & 2.8 & NA & 7.5 \\ \hline
\end{tabular}
}

\caption{\label{Answerwise}Answer-type-wise accuracy (\%) of GPT-4o and Qwen2.5VL [7B] on \datae{} and \datah{} across single-image and multi-image settings. The table highlights the models' performance breakdown by question category (C1-C3) and answer type: (i) Binary, (ii) Numerical, (iii) Categorical, and (iv) Open-ended.}
\end{table*}

\begin{table*}[!ht]
\centering
\begin{tabular}{|l|l|l|l|}
\hline
Country & India & Australia & Pakistan \\ \hline
\multirow{12}{*}{League Teams} & Mumbai Indians & Sydney Thunder & Islamabad United \\
 & Kolkata Knight Riders & Adelaide Strikers & Lahore Qalandars \\
 & Kings XI Punjab & Melbourne Renegades & Karachi Kings \\
 & Royal Challengers Bangalore & Sydney Sixers & Peshawar Zalmi \\
 & Gujarat Lions & Perth Scorchers & Multan Sultans \\
 & Delhi Daredevils & Hobart Hurricanes & Quetta Gladiators \\
 & Sunrisers Hyderabad & Brisbane Heat &  \\
 & Rising Pune Supergiants & Melbourne Stars &  \\

 & Chennai Super Kings &  &  \\
 & Rajasthan Royals &  &  \\
 & Delhi Capitals &  & \\ \hline
\end{tabular}
\caption{List of franchise Cricket teams by Country and League included in the dataset.}
\end{table*}

\subsection{Remaining set of questions and their category}
Table ~\ref{remain_ques} containing the remaining set of questions and their categories.

\begin{table*}[h]
\centering
\resizebox{\textwidth}{!}{
\begin{tabular}{lllcccr}
\toprule
\textbf{Cat.} & \textbf{Category Name} & \textbf{Example Question} & \#sQ's & \#mQ's & \#Total & \multicolumn{1}{c}{\%} \\ \midrule
\multirow{20}{*}{C1} & \multirow{20}{*}{Direct Retrieval \& Simple Inference} & Which bowler has bowled the most no-balls in the match? & 9 & 5 & 14 & 0.93 \\
 &  & Who got out for a duck in the second innings? & 16 & 12 & 28 & 1.87 \\
 &  & Did any batsman score a century in the match? & 25 & 11 & 36 & 2.4 \\
 &  & Which bowler has bowled the maximum maidens? & 4 & 15 & 19 & 1.27 \\
 &  & Did any bowler take a 3-fer in the match? & 24 & 9 & 33 & 2.2 \\
 &  & Did any bowler take a 5-fer in the match? & 18 & 7 & 25 & 1.67 \\
 &  & Did any bowler take a 6-fer in the match? & 23 & 8 & 31 & 2.07 \\
 &  & How many wides were bowled by Team 1? & 27 & 13 & 40 & 2.67 \\
 &  & How many no balls were bowled by Team 1? & 24 & 9 & 33 & 2.2 \\
 &  & How many leg byes did Team 1 concede? & 23 & 6 & 29 & 1.93 \\
 &  & How many byes did Team 1 concede? & 49 & 24 & 73 & 4.87 \\
 &  & How many extras are bowled in match? & 35 & 11 & 46 & 3.07 \\
 &  & Which bowler has bowled the most wides in the match? & 22 & 8 & 30 & 2 \\
 &  & Who got out for a duck in the first innings? & 6 & 5 & 11 & 0.73 \\
 &  & Did any bowler take a 4-fer in the match? & 15 & 7 & 22 & 1.47 \\
 &  & Has {[}Batsman X{]} taken more wickets than {[}Batsman Y{]}? & 30 & 19 & 49 & 3.27 \\
 &  & Which bowler has conceded the most extras? & 57 & 16 & 73 & 4.87 \\
 &  & Who has hit the maximum sixes? & 24 & 8 & 32 & 2.13 \\
 &  & Does {[}Batsmax X{]} hit more sixes than {[}Batsman Y{]}? & 16 & 15 & 31 & 2.07 \\
 &  & How many extras were bowled in the first innings? & 46 & 18 & 64 & 4.27 \\ \midrule
\multirow{11}{*}{C2} & \multirow{11}{*}{Basic Arithmetic Reasoning \& Conditional Logic} & How many batsmen have scored a century? & 19 & 6 & 25 & 1.67 \\
 &  & How many batsmen have been dismissed for a duck? & 20 & 16 & 36 & 2.4 \\
 &  & Which bowler took a 3-fer in the match? & 27 & 21 & 48 & 3.2 \\
 &  & Which bowler took a 5-fer in the match? & - & 9 & 9 & 0.6 \\
 &  & Which bowler took a 6-fer in the match? & - & 2 & 2 & 0.13 \\
 &  & What is {[}Batsman X{]} strike rate? & 46 & 18 & 64 & 4.27 \\
 &  & Did {[}Batsman X{]} score better in the first innings or the second innings? & - & 7 & 7 & 0.47 \\
 &  & Which batsman scored a century in the match? & 11 & 15 & 26 & 1.73 \\
 &  & Which bowler took a 4-fer in the match? & 6 & 11 & 17 & 1.13 \\
 &  & Has {[}Batsman X{]} hit more boundaries than {[}Batsman X{]}? & 13 & 2 & 15 & 1 \\
 &  & Which batsman was dismissed for a golden duck in the match? & 24 & 15 & 39 & 2.6 \\ \midrule
\multirow{10}{*}{C3} & \multirow{10}{*}{Multi-step Reasoning \& Quantitative Analysis} & How many batsmen had a strike rate greater than 70 in the first innings? & 21 & 11 & 32 & 2.13 \\
 &  & Which innings had the maximum maidens? & 4 & 12 & 16 & 1.07 \\
 &  & Has any batsman been dismissed for a golden duck in the match? & 54 & 15 & 69 & 4.6 \\
 &  & Which batsman had the highest strike rate (minimum 10 balls faced)? & 37 & 17 & 54 & 3.6 \\
 &  & Which batsman had the highest boundary percentage? & 35 & 18 & 53 & 3.53 \\
 &  & Which bowler had the better economy rate in the first innings? & 38 & 18 & 56 & 3.73 \\
 &  & Which innings had the higher run rate? & 38 & 15 & 53 & 3.53 \\
 &  & Which batsman had a strike rate greater than 70 in the first innings? & 49 & 13 & 62 & 4.13 \\
 &  & Has the same fielder caught any batsman twice? & 37 & 14 & 51 & 3.4 \\
 &  & Has any batsman been dismissed twice by the same bowler? & 28 & 19 & 47 & 3.13 \\ \midrule
Total &  &  & \multicolumn{1}{l}{1000} & \multicolumn{1}{l}{500} & \multicolumn{1}{l}{1500} & \multicolumn{1}{l}{100} \\ \bottomrule
\end{tabular}
}
\caption{Statistics of single-image and multi-image questions.}
\end{table*}\label{remain_ques}

\subsection{SQL Query for extracting answers}
Table \ref{Question:SQL} contains questions and SQL queries used for getting answers.

\begin{table*}
\centering
\resizebox{\textwidth}{!}
  {
\begin{tabular}{|p{9cm}|p{14cm}|}
\hline
Question & SQL Query \\ \hline
Which bowler has bowled the most wides in the match? & SELECT Bowler\_Name, SUM(WD) AS Total\_Wides FROM bowling GROUP BY Bowler\_Name ORDER BY Total\_Wides DESC LIMIT 1; \\ \hline
Who got out for a duck in the first innings? & SELECT Batsman\_Name FROM batting WHERE Runs = 0 AND Innings = 1 AND `Bowler/Catcher` NOT LIKE 'not out\%'; \\ \hline
Did any bowler take a 4-fer in the match? & SELECT Bowler\_Name, Wicket, Innings FROM bowling WHERE Wicket \textgreater{}= 4; \\ \hline
Has Batsman X taken more wickets than Batsman Y? & SELECT CASE WHEN SUM(CASE WHEN Bowler\_Name = 'Bowler X' THEN Wicket ELSE 0 END) \textgreater SUM(CASE WHEN Bowler\_Name = 'Bowler Y' THEN Wicket ELSE 0 END) THEN 'Yes' ELSE 'No' END AS Result FROM bowling WHERE Bowler\_Name IN ('Bowler X', 'Bowler Y'); \\ \hline
Which bowler has conceded the most extras? & SELECT Bowler\_Name, SUM(WD + NB) AS total\_extras FROM bowling\_data GROUP BY Bowler\_Name ORDER BY total\_extras DESC LIMIT 1; \\ \hline
Who has hit the maximum sixes? & SELECT Batsman\_Name, MAX("6s") AS max\_sixes FROM batting\_data; \\ \hline
Does Batsmax X hit more sixes than Batsman Y? & SELECT Batsman\_Name, SUM(`6s`) AS Total\_Sixes FROM batting WHERE Batsman\_Name IN ('Batsman X', 'Batsman Y') GROUP BY Batsman\_Name; \\ \hline
How many extras were bowled in the first innings? & SELECT SUM(WD + NB) AS Total\_Extras FROM bowling WHERE Innings = 1; for leg bye and bye we calculated manually \\ \hline
Which bowler has bowled the most no-balls in the match? & SELECT Bowler\_Name, SUM(NB) AS Total\_Wides FROM bowling GROUP BY Bowler\_Name ORDER BY Total\_NB DESC LIMIT 1; \\ \hline
Who got out for a duck in the second innings? & SELECT Batsman\_Name FROM batting WHERE Runs = 0 AND Innings = 2 AND `Bowler/Catcher` NOT LIKE 'not out\%'; \\ \hline
Did any batsman score a century in the match? & SELECT Batsman\_Name, Runs, Innings FROM batting WHERE Runs \textgreater{}= 100; \\ \hline
Which bowler has bowled the maximum maidens? & SELECT Bowler\_Name, SUM(Maiden) AS Total\_Maidens FROM bowling GROUP BY Bowler\_Name HAVING Total\_Maidens \textgreater 1 ORDER BY Total\_Maidens DESC; \\ \hline
Did any bowler take a 3-fer in the match? & SELECT Bowler\_Name, Wicket, Innings FROM bowling WHERE Wicket \textgreater{}= 3; \\ \hline
Did any bowler take a 5-fer in the match? & SELECT Bowler\_Name, Wicket, Innings FROM bowling WHERE Wicket \textgreater{}= 5; \\ \hline
Did any bowler take a 6-fer in the match? & SELECT Bowler\_Name, Wicket, Innings FROM bowling WHERE Wicket \textgreater{}= 6; \\ \hline
How many wides were bowled by Team 1? & SELECT SUM(WD) AS Total\_Wides\_By\_Team1 FROM bowling WHERE Innings IN (1, 3); \\ \hline
How many no balls were bowled by Team 1? & SELECT SUM(NB) AS Total\_Wides\_By\_Team1 FROM bowling WHERE Innings IN (1, 3); \\ \hline
How many leg byes did Team 1 concede? & SELECT SUM(byes) AS Total\_Wides\_By\_Team1 FROM bowling WHERE Innings IN (1, 3); \\ \hline
How many byes did Team 1 concede? & SELECT SUM(legbyes) AS Total\_Wides\_By\_Team1 FROM bowling WHERE Innings IN (1, 3); \\ \hline
How many extras are bowled in match? & SELECT SUM(WD + NB) AS Total\_Extras\_In\_Match FROM bowling; \\ \hline
How many batsmen have scored a century? & SELECT COUNT(*) AS Century\_Count FROM batting WHERE Runs \textgreater{}= 100 AND "Bowler/Catcher" NOT LIKE '\%not out\%'; \\ \hline
How many batsmen have been dismissed for a duck? & SELECT CASE WHEN COUNT(*) = 0 THEN 'None' ELSE CAST(COUNT(*) AS TEXT) END AS Duck\_Result FROM batting WHERE Runs = 0 AND "Bowler/Catcher" NOT LIKE '\%not out\%'; \\ \hline
Which bowler took a 3-fer in the match? & SELECT Bowler\_Name, Wicket, Innings FROM bowling WHERE Wicket \textgreater{}= 3; \\ \hline
Which bowler took a 5-fer in the match? & SELECT Bowler\_Name, Wicket, Innings FROM bowling WHERE Wicket \textgreater{}= 5; \\ \hline
Which bowler took a 6-fer in the match? & SELECT Bowler\_Name, Wicket, Innings FROM bowling WHERE Wicket \textgreater{}= 6; \\ \hline
What is Batsman X strike rate? & SELECT ROUND((SUM(Runs) * 100.0 / SUM(Balls)), 2) AS Strike\_Rate FROM batting WHERE Batsman\_Name = 'batsman X' AND Innings = 1; \\ \hline
Did Batsman X score better in the first innings or the second innings? & SELECT CASE WHEN SUM(CASE WHEN Innings = 0 THEN Runs ELSE 0 END) \textgreater SUM(CASE WHEN Innings = 2 THEN Runs ELSE 0 END) THEN '1st Innings' WHEN SUM(CASE WHEN Innings = 2 THEN Runs ELSE 0 END) \textgreater SUM(CASE WHEN Innings = 0 THEN Runs ELSE 0 END) THEN '2nd Innings' ELSE 'None' END AS Better\_Innings FROM batting WHERE Batsman\_Name = 'batsman X'; \\ \hline
Which batsman scored a century in the match? & SELECT Batsman\_Name, Runs, Innings FROM batting WHERE Runs \textgreater{}= 100; \\ \hline
Which bowler took a 4-fer in the match? & SELECT Bowler\_Name, Wicket, Innings FROM bowling WHERE Wicket = 4; \\ \hline
\end{tabular}
}
\caption{Question and its SQL query to extract answer from CSV.}
\label{Question:SQL}
\end{table*}

\begin{table*}[ht!]
\centering
\resizebox{\textwidth}{!}
  {
\begin{tabular}{|p{9cm}|p{14cm}|}
\hline
Question & SQL Query \\ \hline
Has Batsman X hit more boundaries than Batsman X? & SELECT CASE WHEN SUM(CASE WHEN Batsman\_Name = 'batsman X' THEN `4s` + `6s` ELSE 0 END) \textgreater SUM(CASE WHEN Batsman\_Name = 'batsman Y' THEN `4s` + `6s` ELSE 0 END) THEN 'Yes' ELSE 'No' END AS Result FROM batting WHERE Batsman\_Name IN ('batsman X', 'batsman Y'); \\ \hline
Which batsman was dismissed for a golden duck in the match? & SELECT Batsman\_Name FROM batting WHERE Runs = 0 AND `Bowler/Catcher` NOT LIKE 'not out\%'; \\ \hline
How many batsmen had a strike rate greater than 70 in the first innings? & SELECT COUNT(DISTINCT Batsman\_Name) AS Count FROM batting WHERE Innings = 1 AND Strike\_Rate \textgreater 70; \\ \hline
Which innings had the maximum maidens? & SELECT Innings, SUM(Maiden) AS Total\_Maidens FROM bowling GROUP BY Innings HAVING Total\_Maidens \textgreater 1 ORDER BY Total\_Maidens DESC LIMIT 1; \\ \hline
Has any batsman been dismissed for a golden duck in the match? & SELECT Batsman\_Name, Innings FROM batting WHERE Runs = 0 AND Balls = 1; \\ \hline
Which batsman had the highest strike rate (minimum 10 balls faced)? & SELECT Batsman\_Name FROM batting WHERE Innings = 1 AND Balls \textgreater{}= 10 ORDER BY Strike\_Rate DESC LIMIT 1; \\ \hline
Which batsman had the highest boundary percentage? & SELECT Batsman\_Name FROM batting WHERE Innings = 1 AND Balls \textgreater 0 ORDER BY ((({[}4s{]}*4 + {[}6s{]}*6) * 100.0 / Runs)) DESC LIMIT 1; \\ \hline
Which bowler had the better economy rate in the first innings? & SELECT Bowler\_Name, ROUND((SUM(Runs) * 1.0 / SUM(Over)), 2) AS Economy\_Rate FROM bowling WHERE Innings = 1 GROUP BY Bowler\_Name ORDER BY Economy\_Rate ASC LIMIT 1; \\ \hline
Which innings had the higher run rate? & SELECT Innings FROM batting GROUP BY Innings ORDER BY SUM(Runs)*1.0/COUNT(DISTINCT Batsman\_Name) DESC LIMIT 1; \\ \hline
Which batsman had a strike rate greater than 70 in the first innings? & SELECT GROUP\_CONCAT(Batsman\_Name) AS Aggressive\_Batsmen FROM batting WHERE Innings = 1 AND Strike\_Rate \textgreater 70 AND Balls \textgreater{}= 10 GROUP BY Batsman\_Name HAVING Strike\_Rate \textgreater 70; \\ \hline
Has the same fielder caught any batsman twice? & SELECT TRIM(SUBSTR(`Bowler/Catcher`, 3, INSTR(`Bowler/Catcher`, 'b') - 3)) AS Fielder, COUNT(*) AS Catches FROM batting WHERE `Bowler/Catcher` LIKE 'c \%b \%' GROUP BY Fielder HAVING Catches \textgreater 1; \\ \hline
Has any batsman been dismissed twice by the same bowler? & SELECT Batsman\_Name, SUBSTR(`Bowler/Catcher`, INSTR(`Bowler/Catcher`, 'b ') + 2) AS Bowler, COUNT(*) AS Dismissals FROM batting WHERE `Bowler/Catcher` LIKE '\%b \%' GROUP BY Batsman\_Name, Bowler HAVING Dismissals \textgreater 1; \\ \hline
\end{tabular}
}
\caption{Question and its SQL query to extract answer from CSV continued.}
\label{Question:SQL_cont}
\end{table*}

\end{document}